\documentclass[journal,twoside,web]{ieeecolor}

\usepackage{etoolbox}
\makeatletter
\@ifundefined{color@begingroup}%
{\let\color@begingroup\relax
\let\color@endgroup\relax}{}%
\def\fix@ieeecolor@hbox#1{%
\hbox{\color@begingroup#1\color@endgroup}}
\patchcmd\@makecaption{\hbox}{\fix@ieeecolor@hbox}{}{\FAILED}
\patchcmd\@makecaption{\hbox}{\fix@ieeecolor@hbox}{}{\FAILED}
\usepackage{generic}
\usepackage{cite}
\usepackage{amsmath,amssymb,amsfonts}
\usepackage{algorithm,algorithmic}
\usepackage{graphicx}
\usepackage{tikz}
\usepackage{textcomp}
\usepackage{multirow}
\usepackage{array}
\usepackage{makecell}
\usepackage{xcolor}
\usepackage{color}
\usepackage{soul}
\usepackage[export]{adjustbox} 
\usepackage{bm}
\usepackage{tabularx}
\usepackage{booktabs} 
\soulregister\cite7 
\soulregister\cite7 
\soulregister\citet7 
\soulregister\ref7 
\soulregister\pageref7 
\makeatletter
\let\NAT@parse\undefined
\makeatother
\usepackage[colorlinks=true,
           linkcolor=blue,
           anchorcolor=blue,
           citecolor=green,
           urlcolor=blue,
           ]{hyperref}
\def\BibTeX{{\rm B\kern-.05em{\sc i\kern-.025em b}\kern-.08em
    T\kern-.1667em\lower.7ex\hbox{E}\kern-.125emX}}
\begin{document}
\title{USCNet: Transformer-Based Multimodal Fusion with Segmentation Guidance for Urolithiasis Classification}

\author{Changmiao Wang, Songqi Zhang, Yongquan Zhang, Yifei Wang, Liya Liu, Nannan Li, \\ 
Xingzhi Li, Jiexin Pan, Yi Jiang, Xiang Wan, Hai Wang, Ahmed Elazab
\thanks{This work was supported by Guangxi Science and Technology Program (No. FN2504240022), the Guangxi Key R\&D Project (No. AB24010167), Guangdong Basic and Applied Basic Research Foundation (No. 2025A1515011617), Central Funds Guiding the Local Science and Technology Development Project (No. 2025ZYDF106), China Ministry of Education Humanities and Social Sciences Research-Planning Fund Project (No. 23YJA910007), Zhejiang Provincial Philosophy and Social Sciences Planning Project-Major Project (No. 23SYS12ZD), First Class Discipline of Zhejiang - A (Zhejiang University of Finance and Economics-Statistics), and Zhejiang Provincial Bureau of Statistics Statistical Research-Excellent Project (No. 25TJYX09).(Corresponding author: Yongquan Zhang.)}
\thanks{C. Wang and X. Wan are with Shenzhen Research Institute of Big Data, Shenzhen 518172, China (e-mail: wangcm@sribd.cn; wanxiang@sribd.cn).}
\thanks{Y. Zhang, S. Zhang, and Y. Wang are with Zhejiang University of Finance and Economics, Hangzhou 310018, China (e-mail: zyq@zufe.edu.cn; qq1442449344@gmail.com; 220110770228@zufe.edu.cn).}
\thanks{L. Liu is with Anhui University of Finance and Economics, Anhui 233000, China (e-mail: 20231441@aufe.edu.cn).}
\thanks{X. Li, J. Pan, and Y. Jiang are with The Second Affiliated Hospital of Chinese University of Hong Kong (Longgang District People's Hospital of Shenzhen), Shenzhen 518172, China (e-mail: xingzhi02459@gmail.com; jiexinp96@gmail.com; jackyamarelle@gmail.com).}
\thanks{H. Wang is with the School of Engineering and Energy and Harry Butler Institute, Murdoch University, Perth, WA6150, Australia (e-mail: Hai.Wang@murdoch.edu.au).}
\thanks{N. Li is with the School of Computer Science and Engineering, Macau University of Science and Technology, Macao 999078, China (e-mail: nnli@must.edu.mo).}
\thanks{A. Elazab is with Tsinghua Shenzhen International Graduate School, Tsinghua University, Shenzhen 518055, China (e-mail: ahmed.elazab@yahoo.com).}
}

\maketitle

\begin{abstract}
Kidney stone disease ranks among the most prevalent conditions in urology, and understanding the composition of these stones is essential for creating personalized treatment plans and preventing recurrence. Current methods for analyzing kidney stones depend on postoperative specimens, which prevents rapid classification before surgery. To overcome this limitation, we introduce a new approach called the Urinary Stone Segmentation and Classification Network (USCNet). This innovative method allows for precise preoperative classification of kidney stones by integrating Computed Tomography (CT) images with clinical data from Electronic Health Records (EHR). USCNet employs a Transformer-based multimodal fusion framework with CT-EHR attention and segmentation-guided attention modules for accurate classification. Moreover, a dynamic loss function is introduced to effectively balance the dual objectives of segmentation and classification. Experiments on an in-house kidney stone dataset show that USCNet demonstrates outstanding performance across all evaluation metrics, with its classification efficacy significantly surpassing existing mainstream methods. This study presents a promising solution for the precise preoperative classification of kidney stones, offering substantial clinical benefits. The source code has been made publicly available: \href{https://github.com/ZhangSongqi0506/KidneyStone}{https://github.com/ZhangSongqi0506/KidneyStone}.
\end{abstract}


\section{Introduction}
\label{sec:introduction}
\IEEEPARstart{U}{rolithiasis}, a widespread urinary tract disease, impacts millions globally, with its incidence notably increasing in industrialized nations \cite{ref_article20, ref_article27}. This rise is linked to modern lifestyle habits, including poor diet, inadequate hydration, sedentary lifestyle, and diets high in sodium or oxalate. The condition predominantly affects individuals aged 20–40, with a higher occurrence in males, possibly due to the influence of testosterone on stone formation \cite{ref_article4}. Urolithiasis can be divided into infectious and non-infectious types \cite{ref_article30}. Infectious stones, mainly composed of magnesium ammonium phosphate, form in alkaline urine and are often associated with recurrent urinary tract infections caused by urease-producing bacteria \cite{ref_article7}. Conversely, non-infectious stones, such as those formed from calcium oxalate, phosphate, uric acid, and cystine, are primarily due to metabolic disorders, dehydration, or genetic predispositions \cite{ref_article13}, though they can also lead to secondary infections in cases of urinary obstruction or retention. Accurate preoperative prediction of stone composition is vital for selecting appropriate treatment strategies and advising lifestyle changes to prevent recurrence \cite{pearle2014medical}. Traditional diagnostic methods typically depend on postoperative chemical stone analysis\cite{Gay2025} or diagnoses inferred from clinical symptoms and laboratory results \cite{ref_article29}. These methods fall short of providing real-time assessments of stone composition heterogeneity, spatial distribution, and dynamic changes, while overlooking important factors such as stone size, location, and interactions with surrounding tissues. Although computed tomography (CT) imaging is the gold standard for diagnosing urolithiasis \cite{ref_article6, ref_article16, hossain2023vision,sayed2025hybrid}, offering precise stone localization and aiding treatment decisions \cite{xu2024clinical}, it still faces challenges in accurately classifying stone types. Consequently, achieving preoperative classification of in vivo urinary stones using CT imaging remains a significant challenge.

Over the last decade, the rapid progress in deep learning has led to remarkable advancements in the early detection of urinary stones \cite{ref_article15}. Convolutional neural networks (CNNs), particularly those from the ResNet series \cite{ref_proc17}, have significantly enhanced the accuracy of identifying stone features in CT images due to their advanced feature extraction capabilities. ResNet addresses the vanishing gradient issue often encountered in deep network training by incorporating residual connections, which allow for the creation of deeper networks, thereby improving model precision and performance. Several studies have explored the use of CNNs for classifying urinary stones through CT imaging. For instance, Hines \textit{et al.} \cite{ref_article8} and Shen \textit{et al.} \cite{ref_article14} examined CNN applications in the analysis of microscopic CT images to increase the automation and accuracy of stone classification. In related research, Asif \textit{et al.} \cite{ref_article5} introduced StoneNet, which utilizes the MobileNet architecture enhanced with global average pooling, batch normalization, dropout, and dense layers to reduce the number of parameters and increase model robustness. Similarly, Elton \textit{et al.} \cite{ref_article2} developed an automated system for detecting kidney stones and performing volumetric segmentation in non-contrast CT scans, achieving high sensitivity and specificity. Furthermore, Patro \textit{et al.} \cite{ref_article19} proposed a computer-aided diagnostic system that integrates convolution techniques based on the Kronecker product, which effectively minimizes redundancy in feature maps without overlapping convolutions. Cheng \textit{et al.} \cite{ref_article22} introduced ResGANet, incorporating modular group attention blocks to enhance feature representation in medical image classification and segmentation tasks while also reducing parameter complexity. Despite these advancements, relying solely on CT imaging poses significant challenges, particularly in accurately classifying stone types. Many stones, such as calcium oxalate and uric acid stones, have similar radiodensity on CT scans, making differentiation difficult.

In multimodal learning, Baharoon \textit{et al.} proposed the HyMNet model, which enhances feature extraction and disease prediction accuracy by combining visual data from fundus images with clinical information on cardiovascular metabolic risk factors \cite{ref_article24}. However, its reliance on simple feature concatenation limits its ability to fully explore deep correlations and interactions between modalities, potentially leading to redundant or lost information and affecting performance on complex datasets. To improve multimodal fusion, a Transformer-based mechanism can be introduced. Recent studies have demonstrated the effectiveness of this approach in medical applications: Yu \textit{et al.} \cite{yu2025prnet} developed ICH-PRNet, a novel cross-modal network for ICH prognosis prediction that effectively combines CT scans and clinical notes through an advanced joint-attention mechanism, while Yu \textit{et al.} \cite{yu2024scnet} created ICH-SCNet, an innovative framework leveraging CLIP-guided SAM technology for simultaneous segmentation and classification. The Transformer architecture \cite{ref_article11} revolutionized deep learning by replacing traditional CNNs with self-attention, capturing global dependencies among sequence elements. Initially designed for textual data, the Transformer's applicability to other data types was limited due to its focus on discrete word sequences. To extend its use, Google Research developed the Vision Transformer (ViT) \cite{ref_article12}, which divides images into fixed-size patches and processes them as sequences through a Transformer encoder, opening new possibilities for multimodal data processing. Additionally, Hatamizadeh \textit{et al.} \cite{ref_proc18}proposed the UNETR model, which combines the Transformers' global context modeling with U-Net's \cite{ronneberger2015u} precise localization, achieving superior performance in medical image segmentation. Recently, Xing \textit{et al.} \cite{xing2024hybrid} introduced a hybrid masked image modeling method that significantly enhances 3D medical image segmentation accuracy. 

Current methods for integrating segmentation with classification tasks fall into two main categories: segmentation-assisted classification and segmentation-classification multi-task frameworks. In the segmentation-assisted classification approach, Zhu \textit{et al.} \cite{ref_proc3} introduced the SegPrompt model, which utilizes pretrained models that employ segmentation maps as prompts for classifying kidney stones. However, this method has several drawbacks: it does not incorporate segmentation during training, the segmentation maps are not directly linked to classification, and it depends heavily on pretraining. On the other hand, in the segmentation-classification multi-task framework, Saeed \textit{et al.} \cite{ref_article23} developed the TMSS network, which conducts cancer segmentation and survival prediction simultaneously. Although TMSS uses Transformers for comprehensive processing of multimodal data, it treats segmentation and classification as independent tasks without sharing features interactively. Furthermore, its fixed weighting mechanism for loss calculation restricts dynamic adjustments, leading to task imbalances that adversely affect overall performance.

Building on prior research, the accurate preoperative classification of urinary stones presents three significant challenges. First, single CT imaging falls short in capturing the patient-specific clinical context that characterizes infectious stones. Second, current segmentation-assisted methods only offer static spatial prompts, lacking the dynamic interaction necessary between classification and segmentation. Third, multi-task frameworks encounter inherent conflicts between the optimization objectives for segmentation and classification. To address these challenges, we propose a novel model, USCNet, designed for in vivo urinary stone classification. USCNet integrates Transformer-based multimodal fusion with segmentation-assisted classification, leveraging multimodal data from CT images and Electronic Health Records (EHR) \cite{wang2023ehr}. As a multi-task multimodal framework, USCNet simultaneously performs segmentation and classification, enhancing the synergy between these processes. By incorporating clinical data and segmentation feature maps into the classification workflow via self-attention mechanisms, the model ensures improved accuracy and contextual understanding. The key contributions of USCNet are as follows:

\begin{figure*}[!tbp]
\centering
\includegraphics[width=7.16in, height=3 in]{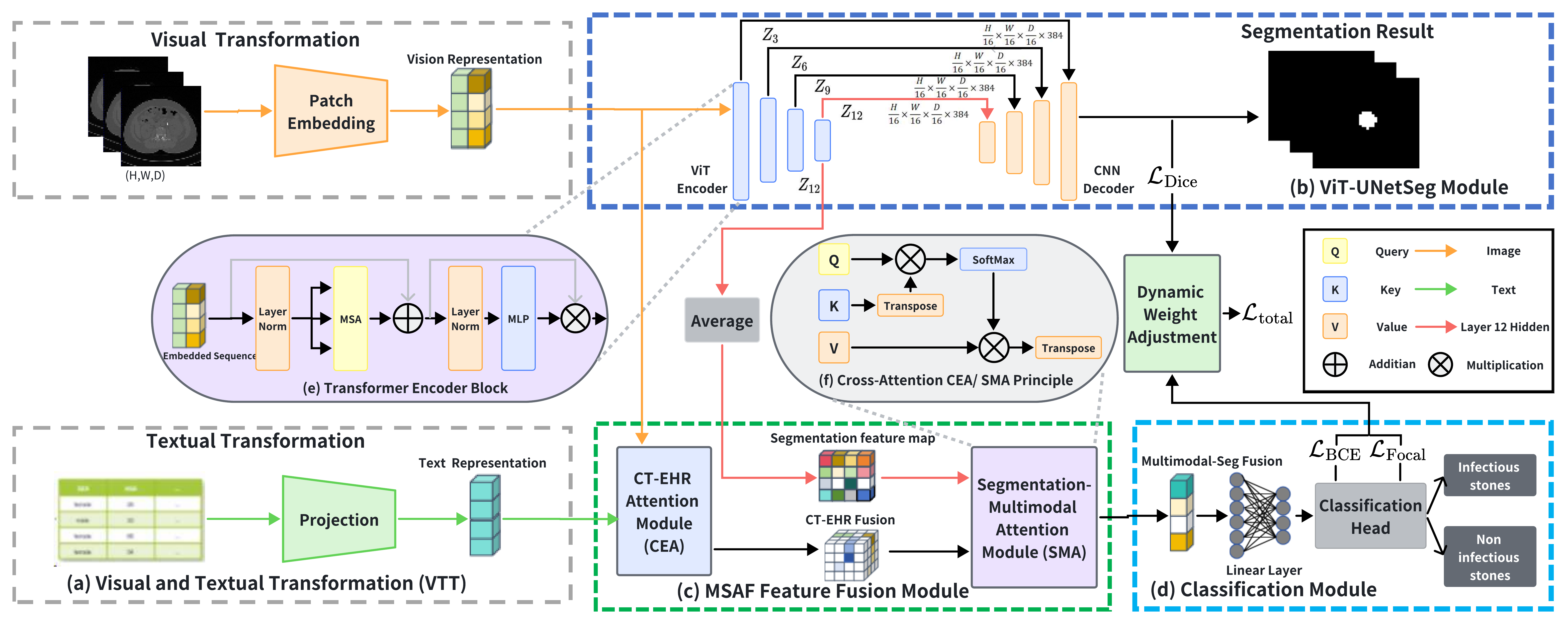}
\caption{The architecture of the USCNet: (a) the visual and textual transformation module, (b) the ViT-UNetSeg module, (c) the MSAF feature fusion module, (d) the classification module,  (e) the Transformer encoder block, and (f) the cross-attention principle.}
\label{architecture}
\end{figure*}

1) To overcome the limitations of traditional unimodal methods, we developed the CT-EHR attention module, which enhances the fusion of imaging features with clinical context. This module dynamically integrates CT images with patient-specific clinical data at the feature level through cross-modal interaction mechanisms. By incorporating clinical background information, the model can more effectively distinguish infectious stones.

2) Unlike SegPrompt \cite{ref_proc3}, which utilizes segmentation maps as static prompts, we introduce the Segmentation Multimodal Attention (SMA) module. The primary aim of this module is to enable the rich spatial information generated during segmentation to actively and dynamically inform classification decisions. By employing trainable attention mechanisms, the SMA module autonomously establishes semantic relationships between segmentation features and classification tasks, ensuring the model concentrates on the most discriminative anatomical structures for differentiating stone types.

3) In our segmentation-assisted classification framework, dependable early-stage segmentation features are crucial for effective classification outcomes. Unlike existing multi-task frameworks \cite{ref_article23, ref_article24} that rely on fixed-weight loss functions, we propose a dynamic loss function. This function adaptively adjusts the loss weights by continuously monitoring the performance metrics of each task in real-time. This approach allows for the coordinated optimization of both segmentation and classification tasks during training, offering a novel optimization paradigm for multi-task learning in complex scenarios.

The remainder of this paper is organized as follows: Section \ref{Methods} details the proposed USCNet architecture, including the Visual and Textual Transformation Module, ViT-UNetSeg Module, and Multimodal-Segmentation Attention Fusion Module, with emphasis on the dynamic weight adjustment for total loss. Section \ref{Experiments} presents the comprehensive experimental study, covering: (1) data preparation and implementation details, (2) comparative experiments with state-of-the-art methods, and (3) ablation studies investigating key modules, hierarchical feature combinations, loss components, and adaptive loss functions. Finally, Section \ref{Conclusion} concludes the study.



\section{Methods}
\label{Methods}
The network architecture is effectively designed, as shown in Fig. \ref{architecture}, and includes four main components: the Visual and Textual Transformation (VTT) module (Section \ref{VTT}), the segmentation module, the Multimodal-Segmentation Attention Fusion (MSAF) module (Section \ref{MSAF}), and the classification module. The VTT module preprocesses and transforms CT images and EHR data into a unified feature space by embedding CT images and projecting EHR data for a harmonized representation. The segmentation module (Section \ref{Seg}) utilizes a UNETR \cite{ref_proc18} structure with a ViT \cite{ref_article12} encoder to segment the visual representation output from the VTT module, thereby obtaining segmentation results. Our primary objective is to assess the infectivity of stones, and the Transformer module within the UNETR \cite{ref_proc18} architecture excels at pinpointing the spatial locations of stones in images. The MSAF module comprises the CEA and SMA modules, which integrate and refine features by selectively emphasizing clinically relevant information derived from images and EHR data. The classification module primarily includes the MSAF module, a linear layer, and a classification head. Its main function is to derive classification results from the MSAF module's output via the linear layer and classification head. Furthermore, the overall loss function for segmentation and classification is dynamically adjusted by the loss weighting module (Section \ref{DC}) to ensure balanced optimization of both tasks.

\subsection{Visual and Textual Transformation Module}\label{VTT}
In this module, CT images undergo preprocessing through the Visual Transformation process. Specifically, the 3D image with dimensions \( \boldsymbol{x} \in \mathbb{R}^{\boldsymbol{H} \times \boldsymbol{W} \times \boldsymbol{D}} \) is reshaped into a sequence of flat 2D patches \( \boldsymbol{x}_p \in \mathbb{R}^{\boldsymbol{N} \times (\boldsymbol{P}^3)} \). Here, \(\boldsymbol{W}\), \(\boldsymbol{H}\), and \(\boldsymbol{D}\) denote the width, height, and depth of the CT images, respectively, while \(\boldsymbol{P}\) signifies the dimension of each patch. The total number of patches \(\boldsymbol{N}\) is calculated as follows:

\begin{equation}
\begin{split}
\label{patch_formula}
    \boldsymbol{N} = \frac{\boldsymbol{W} \times \boldsymbol{H} \times \boldsymbol{D}}{\boldsymbol{P}^3},
\end{split}
\end{equation}
where \(\boldsymbol{P} \times \boldsymbol{P} \times \boldsymbol{P}\) represents the dimensions of each patch. To preserve the spatial relationships among segments, position embedding \cite{ref_article12} is added to each patch after partitioning.

Simultaneously, the EHR data is transformed into a comparable feature space through the Textual Transformation. This process results in a linear feature representation with dimensions \([\boldsymbol{B}, \boldsymbol{E}]\), where \(\boldsymbol{B}\) and \(\boldsymbol{E}\) represent the batch and embedding sizes, respectively.The primary aim of this module is to harmonize heterogeneous data modalities, specifically high-dimensional CT images and structured EHR text, into a unified format. This harmonization is crucial for facilitating seamless integration and analysis in the subsequent phases of multimodal analysis. This preparatory step lays the foundation for a comprehensive cross-modal examination later in the multimodal analytical workflow.

\textbf{Remark 1}. The Visual and Textual Transformation Module bridges the modality gap between medical imaging and clinical data through innovative 3D patch embedding and synchronized positional encoding, preserving critical anatomical relationships while enabling direct cross-modal feature alignment.

\subsection{The  ViT-UNetSeg Module}\label{Seg}
The segmentation module is tailored for 3D medical image segmentation, integrating a ViT-based encoder with a U-Net style decoder to extract spatial features at multiple scales and produce accurate segmentation masks. The encoder initially divides the 3D input volume into non-overlapping patches of size \(\boldsymbol{P} \times \boldsymbol{P} \times \boldsymbol{P}\), where \(\boldsymbol{P} = 16\). These patches are then flattened and processed through 12 Transformer layers, each possessing a hidden size of 384, to capture hierarchical global contextual information. Intermediate features from the encoder's layers \(\mathbf{Z}_3\), \(\mathbf{Z}_6\), \(\mathbf{Z}_9\), and \(\mathbf{Z}_{12}\) (the final layer) are extracted to represent multi-scale spatial information.

The decoder employs a hierarchical upsampling structure, consisting of multiple UnetrUpBlock and UnetrPrUpBlock layers \cite{ref_proc18}. Each upsampling block incrementally increases the spatial resolution of the feature maps while reducing the number of channels. Skip connections are employed to merge multi-scale features from the encoder (\(\mathbf{Z}_3\), \(\mathbf{Z}_6\), \(\mathbf{Z}_9\), and \(\mathbf{Z}_{12}\)) with the corresponding decoder layers. This approach ensures the effective utilization of both local details and global context for precise 3D segmentation. Finally, the UnetOutBlock \cite{ref_proc18} generates the 3D segmentation mask by transforming the high-resolution feature maps into the required number of output channels. This modular design enables the model to harness the global feature extraction strengths of the Transformer along with the local feature refinement capabilities of the U-Net.

\textbf{Remark 2}. This segmentation module innovatively combines a ViT encoder with a U-Net decoder through 3D patch embedding and hierarchical Transformer features. The progressive upsampling with skip connections achieves synergistic optimization of global context modeling and local detail preservation, inheriting both the Transformer's global perception and the U-Net's local refinement capabilities.

\subsection{Multimodal-Segmentation Attention Fusion Module }\label{MSAF}
The MSAF module comprises CEA and SMA modules, arranged sequentially and operating on a cross-attention mechanism. Initially, the CEA module combines the visual and textual representations from the VTT module to create the CT-EHR Fusion Representation (CEFR). Subsequently, the SMA module merges the segmentation feature map, sourced from the 12th transformer block of the ViT encoder, with the CEFR to generate the Multimodal-Segmentation Fusion Representation (MSFR). This MSFR output is then used for classification.

\textbf{CT and EHR Attention:}
Within the MSAF module, the CEA submodule plays a crucial role in integrating CT image features with EHR data. It uses the vision representation as queries (\(\boldsymbol{Q}_1\)) and the EHR data as keys (\(\boldsymbol{K}_1\)) and values (\(\boldsymbol{V}_1\)). Based on the attention mechanism described in \cite{ref_article11}, the CEFR is computed as follows:

\begin{equation}
\label{cea_formula}
    \text{Attention}(\boldsymbol{Q}_1, \boldsymbol{K}_1, \boldsymbol{V}_1) = \text{softmax}\left(\frac{\boldsymbol{Q}_1\boldsymbol{K}_1^T}{\sqrt{\boldsymbol{d}_{k_1}}}\right)\boldsymbol{V}_1,
\end{equation}
where \(\boldsymbol{d}_{k_1}\) is the dimension of \(\boldsymbol{K}_1\). This setup allows the model to focus on image patches informed by EHR data, ensuring that the analysis is enriched with patient-specific context. \par

\textbf{Segmentation Multimodal Attention:}
The SMA submodule enhances cross-modal data integration by combining the segmentation feature map with CEFR. It uses the segmentation feature map as queries (\(\boldsymbol{Q}_2\)) and the CEFR as keys (\(\boldsymbol{K}_2\)) and values (\(\boldsymbol{V}_2\)). The MSFR is calculated as follows:

\begin{equation}
\label{sma_formula}
    \text{Attention}(\boldsymbol{Q}_2, \boldsymbol{K}_2, \boldsymbol{V}_2) = \text{Softmax}\left(\frac{\boldsymbol{Q}_2\boldsymbol{K}_2^T}{\sqrt{\boldsymbol{d}_{k_2}}}\right)\boldsymbol{V}_2,
\end{equation}
where \(\boldsymbol{d}_{k_2}\) is the dimension of \(\boldsymbol{K}_2\). This fusion process enhances the model's ability to focus on significant areas of the image identified by segmentation features. Consequently, the SMA ensures that tasks such as classification or subsequent analyses prioritize the most relevant regions of the image, informed by both segmentation and the broader multimodal context.

\textbf{Remark 3}. The MSAF module achieves triple fusion of imaging, clinical data, and anatomical segmentation through cascaded CEA and SMA attention mechanisms. The CEA submodule establishes patient-specific context by treating CT features as queries and EHR as keys and values, while the SMA submodule further guides multimodal representation to focus on critical regions using segmentation features as queries, forming a hierarchical cross-modal attention architecture.

\subsection{Dynamic Weight Adjustment for Total Loss}\label{DC}
The USCNet architecture employs a composite loss function specifically designed for its dual tasks of segmentation and classification. This integrated loss function consists of three components: Dice loss \cite{ref_proc9} for segmentation, and binary cross-entropy (BCE) loss coupled with focal loss \cite{ref_proc10} for classification. The overall loss is calculated as a dynamically weighted sum of these components, facilitated by an innovative automatic optimization strategy. This approach ensures a balanced focus between segmentation and classification tasks, with adjustments based on their respective performance metrics.

\textbf{Dice Loss} is used for the segmentation task, and its formula is defined as:
\begin{equation}
\label{dice_loss}
    \mathcal{L}_{\text{Dice}} = 1 - \frac{2 \sum_{i} p_i g_i}{\sum_{i} p_i + \sum_{i} g_i},
\end{equation}
where \(p_i\) represents the predicted value and \(g_i\) is the ground truth.

\textbf{BCE Loss} addresses the classification task, computed as:
\begin{equation}
\label{bce_loss}
    \mathcal{L}_{\text{BCE}} = -\frac{1}{\boldsymbol{N}} \sum_{i} \left[ g_i \log(p_i) + (1 - g_i) \log(1 - p_i) \right],
\end{equation}
where \(p_i\) is the predicted probability, \(g_i\) is the ground truth label, and \(N\) is the number of samples.

\textbf{Focal Loss} is employed to tackle class imbalance in classification, defined as:
\begin{equation}
\label{focal_loss}
    \mathcal{L}_{\text{Focal}} = -\frac{1}{\boldsymbol{N}} \sum_{i} \left[ g_i (1 - p_i)^\gamma \log(p_i) + (1 - g_i) p_i^\gamma \log(1 - p_i) \right],
\end{equation}
where \(\gamma\) is the focusing parameter that adjusts the weight of hard and easy samples.

The total loss function is expressed as:
\begin{flalign}&&
    \mathcal{L}_{\text{total}} = \omega_{\text{Dice}} \mathcal{L}_{\text{Dice}} + \omega_{\text{Class}} (\lambda \mathcal{L}_{\text{BCE}} + (1 - \lambda) \mathcal{L}_{\text{Focal}}),&&
\end{flalign}
where \(\omega_{\text{Dice}}\) and \(\omega_{\text{Class}}\) denote the weights assigned to the Dice loss and the composite classification losses, respectively. \(\lambda\) is a predetermined constant balancing the BCE and focal losses, such as a 1:10 ratio.

\textbf{Weight Adjustment Strategy:}
As shown in Algorithm \ref{alg1}, when the Dice score \(S_{\text{Dice}}\) is 0.8 or lower, the weights are dynamically adjusted: \(\omega_{\text{Dice}} = 1 - S_{\text{Dice}}\) and \(\omega_{\text{Class}} = S_{\text{Dice}}\). The classification loss weights are divided into \(\lambda \mathcal{L}_{\text{BCE}}\) and \((1 - \lambda) \mathcal{L}_{\text{Focal}}\). When the Dice score exceeds 0.8, fixed weights prioritize classification: \(\omega_{\text{Dice}} = 0.2\) and \(\omega_{\text{Class}} = 0.8\), maintaining the division of classification loss components.

\begin{algorithm}[H]
\caption{Dynamic weight adjustment for USCNet.}\label{alg:alg1}
\begin{algorithmic}
\STATE Initialize $\omega_{\text{Dice}} = 0.8$, $\lambda = 0.1$, \COMMENT{$\omega_{\text{Class}} = \omega_{\text{BCE}} + \omega_{\text{Focal}}$}
\FOR{each training epoch}
    \STATE Compute Dice score $S_{\text{Dice}}$ on validation set
    \IF{$S_{\text{Dice}} \leq 0.8$}
        \STATE $\omega_{\text{Dice}} \gets 1 - S_{\text{Dice}}$
        \STATE $\omega_{\text{BCE}} \gets \lambda S_{\text{Dice}}$ 
        \STATE $\omega_{\text{Focal}} \gets S_{\text{Dice}} - \omega_{\text{BCE}}$ \COMMENT{$(1-\lambda)S_{\text{Dice}}$}
    \ELSE
        \STATE $\omega_{\text{Dice}} \gets 0.2$
        \STATE $\omega_{\text{BCE}} \gets \lambda \times 0.8$ \COMMENT{$0.08$}
        \STATE $\omega_{\text{Focal}} \gets 0.8 - \omega_{\text{BCE}}$ \COMMENT{$0.72$}
    \ENDIF
    \STATE $\mathcal{L}_{\text{total}} = \omega_{\text{Dice}}\mathcal{L}_{\text{Dice}} + \omega_{\text{BCE}}\mathcal{L}_{\text{BCE}} + \omega_{\text{Focal}}\mathcal{L}_{\text{Focal}}$
\ENDFOR
\end{algorithmic}
\label{alg1}
\end{algorithm}

This dynamic weight adjustment strategy enables the model to shift focus towards enhancing classification performance as segmentation accuracy improves. Initially, a higher segmentation loss weight encourages rapid segmentation convergence. As segmentation accuracy improves, the focus shifts to classification. By recalibrating these dynamic weights during each training epoch, the model maintains a balanced optimization of both tasks, significantly enhancing overall efficacy.

\textbf{Remark 4}. This module proposes a dynamic task-balancing strategy that intelligently adjusts segmentation and classification loss weights based on Dice scores. It prioritizes segmentation convergence initially, then automatically shifts focus to classification when reaching the threshold, establishing a "segmentation-first-classification-later" progressive training paradigm. This design resolves optimization conflicts in multi-task learning inherent to fixed-weight approaches.

\begin{table*}[!t] 
\small
\centering
\caption{Comparisons with SOTA methods (best in bold, second-best underlined).}
\label{T1}
\begin{adjustbox}{width=1.05\textwidth, center}
\setlength{\tabcolsep}{1.2mm}
\renewcommand{\arraystretch}{1.3}
\begin{tabular}{c|c|ccccc|cccc}
\hline
\multicolumn{1}{c|}{\multirow{2}{*}{No.}} & \multicolumn{1}{c|}{\multirow{2}{*}{Methods}} & \multicolumn{5}{c|}{Classification}  & \multicolumn{3}{c}{Segmentation} \\ 
\cline{3-10} &\multicolumn{1}{c|}{} & \multicolumn{1}{c}{Acc$\uparrow$(\%)} & \multicolumn{1}{c}{F1$\uparrow$(\%)} & \multicolumn{1}{c}{Rec$\uparrow$(\%)} & \multicolumn{1}{c}{Pre$\uparrow$(\%)} & \multicolumn{1}{c|}{AUC$\uparrow$(\%)} & \multicolumn{1}{c}{Dice$\uparrow$(\%)} & \multicolumn{1}{c}{IoU$\uparrow$(\%)} & \multicolumn{1}{c}{HD95$\downarrow$(mm)} \\ 
\hline
1&ResNet34 \cite{ref_proc17} & 56.25$\pm$2.14 & 10.43$\pm$3.51 & 5.50$\pm$2.82 & 44.90$\pm$3.21 & 52.75$\pm$2.30 & - & - & - \\
2&ResNet50 \cite{ref_proc17} & 62.13$\pm$1.82 & 28.96$\pm$2.92 & 19.26$\pm$2.55 & 58.33$\pm$2.79 & 55.03$\pm$2.12 & - & - & - \\
3&ResNet101 \cite{ref_proc17} & 63.97$\pm$1.72 & 38.75$\pm$2.64 & 28.44$\pm$2.34 & 60.47$\pm$2.42 & 58.05$\pm$1.91 & - & - & - \\
4&StoneNet \cite{ref_article5} & 59.93$\pm$2.01 & 57.22$\pm$2.25 & 49.69$\pm$2.64 & 40.07$\pm$3.52 & 50.04$\pm$2.46 & - & - & - \\
5&SegPrompt \cite{ref_proc3} & 61.27$\pm$1.96 & 46.94$\pm$2.45 & 42.20$\pm$2.40 & 52.87$\pm$2.81 & 58.52$\pm$2.01 & - & - & - \\ \hline
6&nnUnet \cite{ref_article21} & - & - & - & - & - & \textbf{86.35$\pm$0.41} & \textbf{75.68$\pm$0.41} & \textbf{3.50$\pm$0.35} \\ 
7&UNETR \cite{ref_proc18} & - & - & - & - & - & 82.27$\pm$1.02 & 69.18$\pm$1.25 & 3.88$\pm$0.51 \\ 
\hline
8&ResGANet \cite{ref_article22} & 62.50$\pm$2.02 & 47.17$\pm$2.82 & 44.31$\pm$2.62 & 46.96$\pm$3.11 & 56.37$\pm$2.36 & 83.85$\pm$1.22 & 72.08$\pm$1.22 & 4.80$\pm$0.45 \\ 
9&TMSS \cite{ref_article23} & 74.63$\pm$1.84 & 57.67$\pm$2.52 & 43.12$\pm$2.85 & \underline{87.04$\pm$2.22} & 81.54$\pm$1.73 & 79.13$\pm$1.43 & 65.43$\pm$1.44 & 6.20$\pm$0.60 \\ 
10&HyMNet \cite{ref_article24} & \underline{85.66$\pm$1.55} & \underline{82.81$\pm$1.95} & \underline{86.23$\pm$1.46} & 79.66$\pm$2.53 & \underline{85.76$\pm$1.35} & 77.14$\pm$1.52 & 62.67$\pm$1.55 & 5.50$\pm$0.75 \\
\hline
\textbf{$\star$} & \textbf{USCNet (Ours)} & 
\textbf{91.91$\pm$1.42} & 
\textbf{90.00$\pm$1.74} & 
\textbf{90.83$\pm$0.94} & 
\textbf{89.19$\pm$3.13} & 
\textbf{91.73$\pm$1.30} & 
\underline{83.98$\pm$1.09} & 
\underline{72.12$\pm$1.42} & 
\underline{3.73$\pm$0.82} \\ 
\hline
\end{tabular}
\end{adjustbox}
\end{table*}

\section{Experiments}
\label{Experiments}
\subsection{Data Preparation and Preprocessing}\label{3.1}
We collected data from 642 patients who underwent CT imaging followed by surgical treatment for urolithiasis. This dataset includes preoperative urological CT images in DICOM format, preoperative complete blood count results, urinalysis, and postoperative stone composition analysis. Overall, the dataset comprises 1,355 CT images of kidney stones. The postoperative stone composition analysis, regarded as the gold standard, was used to classify the stones into infectious and non-infectious categories. Stone masks were annotated by four experienced urologists using 3D Slicer software (version 5.2.2) based on surgical outcomes, with the accuracy of these annotations validated by two experienced radiologists.

This study utilized seven key clinical variables extracted from preoperative EHR: \textbf{1) Age, 2) Gender, 3) Blood Leukocyte Count, 4) Serum Creatinine, 5) Urine Leukocyte Count, 6) Urine pH,} and \textbf{7) Stone Location}. These variables were chosen based on their clinical significance and availability in routine preoperative evaluations. Notably, all variables, including stone location derived from CT imaging, were collected exclusively during the preoperative period. The ground truth labels, infectious versus non-infectious stones, were determined through postoperative composition analysis. This distinction between preoperative features and postoperative labels was integral to the study's design, ensuring a clear temporal separation that effectively mitigates the risk of data leakage. To further enhance the dataset's integrity, variables with substantial missing data or those not routinely assessed during preoperative care, such as urine culture results, were deliberately excluded to minimize selection bias and maintain consistency in clinical applicability.

The CT images were initially resampled to a uniform resolution of 1 mm³ to ensure consistency across all data. Given that a single patient could have multiple stones, images were cropped to isolate each stone region. First, a region of interest (ROI) was identified to locate each stone. The stones were then categorized as infectious or non-infectious, and these regions were extracted into cubes measuring 48 mm on each side. Only cubes containing stones with a volume of at least 0.06 mm³ were selected for further analysis, resulting in a total of 2,113 stone image cases.

To ensure robust evaluation, the dataset was divided using a 5-fold cross-validation approach. Before being input into the network, the CT images underwent window leveling adjustment to the range of (-400, 2000) Hounsfield units, followed by normalization. Meanwhile, continuous variables in the clinical dataset were standardized, and categorical variables were converted into a one-hot encoded format.

\subsection{Implementation Details}
Our experiments were conducted on a 48G NVIDIA A40 GPU running CUDA Version 12.2, using Python 3.11 and PyTorch 2.4. 
Using an input size of \(48^3\) voxels and a batch size of 128, the training process consumed 11.52 GB of GPU memory at its peak. Each training fold required 0.92 hours to complete, equivalent to 0.46 GPU-hours.
To ensure robust model evaluation, we employed a five-fold cross-validation strategy. This involved dividing the dataset into five subsets, with each subset serving once as the validation set while the remaining subsets were used for training. We maintained consistent class distributions across all folds through stratified splitting.

The model was trained for 200 epochs with a batch size of 128 and an initial learning rate of 0.001. We utilized the Adam optimizer, configured with a weight decay of 0.001 and betas set to (0.9, 0.99). These parameters controlled the exponential decay rates for the first and second moment estimates, aiding in momentum handling and adaptive learning rate adjustments. To enhance training stability, we implemented the ReduceLROnPlateau learning rate scheduler, which reduces the learning rate when the validation loss plateaus. The scheduler was set with a reduction factor of 0.1 and a patience of 10 epochs, ensuring adjustments only when necessary.

The segmentation module employed a ViT architecture based on the UNETR framework \cite{ref_proc18}, consisting of 12 layers with 12 attention heads each. The adaptive loss function began with initial weights of 0.2 for the classification loss and 0.8 for the Dice loss. We set \(\lambda=0.1\), resulting in weights of 0.02 for the BCE loss and 0.18 for the focal loss, which were adjusted based on segmentation performance. To address class imbalance, the focal loss was modulated by parameters \(\gamma\) and \(\alpha\), set at 2 and 0.25, respectively, to focus on hard-to-classify samples and balance positive and negative sample weights. This configuration was designed to enhance model performance on imbalanced datasets.

\subsection{Comparative Experiments}\label{3.3}
\subsubsection{Comparative Methods}
We evaluated the proposed USCNet model against several traditional ResNet architectures \cite{ref_proc17}, including ResNet34 \cite{ref_proc17}, ResNet50 \cite{ref_proc17}, and ResNet101 \cite{ref_proc17}, to assess its classification performance. Additionally, we compared USCNet with StoneNet \cite{ref_article5} and SegPrompt \cite{ref_proc3}, two prominent models in the field of stone classification. These benchmark models were used to demonstrate USCNet's effectiveness in classification tasks.

Furthermore, we assessed USCNet's segmentation capabilities by comparing it with nnU-net \cite{ref_article21}, a well-regarded automated medical image segmentation framework based on U-Net, known for its strong generalization and efficiency in medical image segmentation. We also included UNETR \cite{ref_proc18}, an advanced Transformer-based benchmark model for medical image segmentation, to provide comparative performance analysis. Furthermore, we conducted comprehensive comparisons with ResGANet \cite{ref_article22}, TMSS \cite{ref_article23}, and HyMNet \cite{ref_article24} in terms of both classification and segmentation performance. Notably, ResGANet employs independent modules for classification and segmentation without exploiting potential inter-task relationships, while both TMSS and HyMNet are designed as multi-task, multi-modal fusion models that can simultaneously handle classification and segmentation tasks through multi-modal data integration. It should be noted that since HyMNet was originally designed for survival prediction tasks, we adapted its output layer structure to accommodate classification tasks. Through these comparative experiments, we have comprehensively demonstrated USCNet's superior performance in jointly handling both classification and segmentation tasks. 

\subsubsection{Evaluation Metrics}
We employed five key evaluation metrics to assess the classification performance of our proposed model in comparison to state-of-the-art approaches: Accuracy (Acc), F1 Score (F1), Recall (Rec), Precision (Pre), and Area Under the Curve (AUC). While Acc measures the overall proportion of correct predictions, it can be misleading in imbalanced datasets. In such cases, Pre and Rec are particularly valuable, as Pre evaluates the Acc of positive predictions, and Rec assesses the model's ability to identify all actual positives. The F1, which balances Pre and Rec, is especially useful for optimizing performance in imbalanced settings.

To evaluate segmentation accuracy, we employed three key metrics: the Dice Coefficient (Dice), Intersection over Union (IoU), and the 95th percentile Hausdorff Distance (HD95). The Dice Coefficient and IoU assess the overlap between predicted and actual segmentation masks, with higher values reflecting improved performance. The HD95 measures the boundary distance between the predicted and ground truth contours, where lower values signify greater boundary delineation accuracy. Collectively, these metrics provide a robust and comprehensive framework for assessing model performance across both segmentation and classification tasks.

\begin{figure}[!t]
\centerline{\includegraphics[width=\columnwidth]{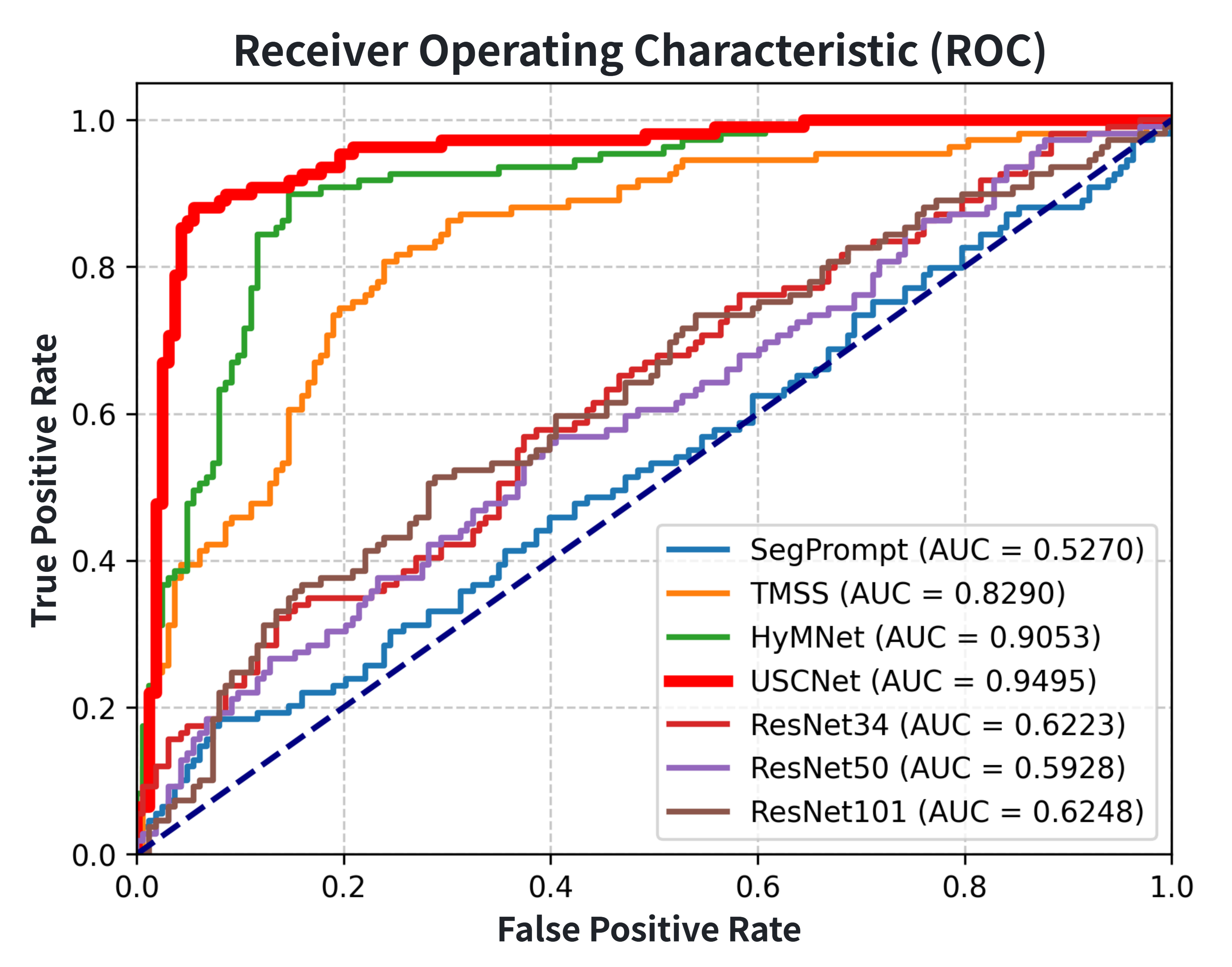}}
\caption{ROC curve comparison of diagnostic accuracy between USCNet and established methods for infectious urolithiasis classification. }
\label{ROC}
\end{figure}

\subsubsection{Comparative Results}
The results, presented in Table \ref{T1}, demonstrate that USCNet surpasses all comparative models in classification tasks across various metrics. Among the ResNet variants, the highest accuracy observed was 63.97\%, with ResNet101 achieving the best performance in this family. StoneNet achieved an Acc of 59.93\%, similar to the ResNet models, whereas SegPrompt reported an Acc of 61.27\%. ResGANet, on the other hand, achieved 62.50\% Acc in classification tasks. These relatively low Acc rates stem from the unique challenges associated with in vivo stone classification. The subtle variations in stone shape and size, coupled with the absence of color information in CT scans, hinder feature extraction and classification Acc. Consequently, single-modal models like StoneNet and SegPrompt, designed for in vitro stone classification, struggle with in vivo tasks due to their reliance solely on CT images.

\begin{figure}[!tbp]
    \centering
    
    \hspace*{0.1\linewidth}\includegraphics[width=0.9\linewidth]{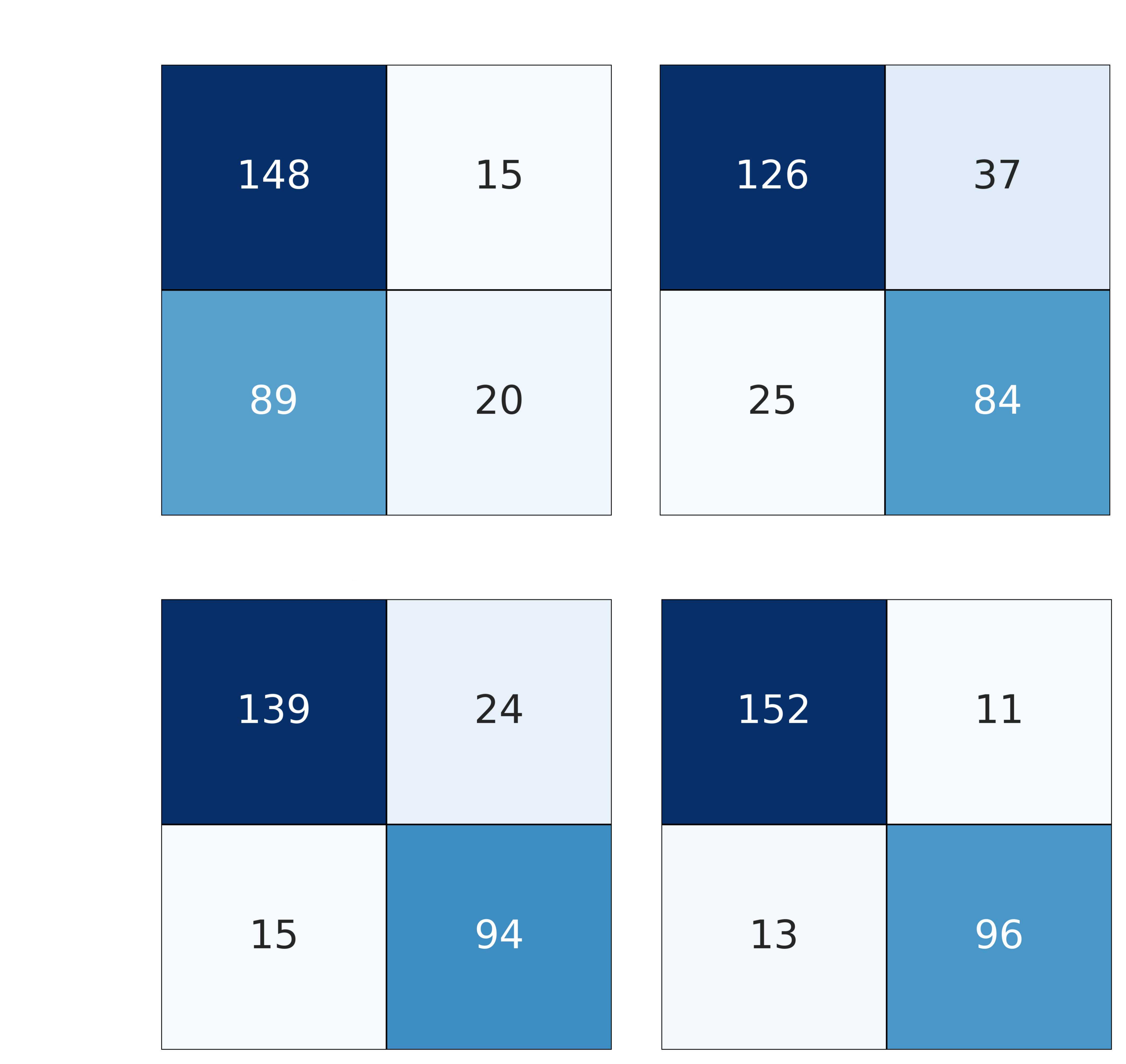}
    
    \begin{tikzpicture}[overlay]
        \node[anchor=north] at (0.05\linewidth, 2mm) {\textbf{Predicted}};
        \node[anchor=north] at (0.27\linewidth, 6mm) {
            \begin{tabular}{cccc}
                Non-infectious & Infectious  \\
            \end{tabular}   
        };
        \node[anchor=north] at (0.09\linewidth, 81mm) {
            \begin{tabular}{c@{\hspace{21mm}}c}
                SegPrompt & TMSS  \\
            \end{tabular}   
        };
        \node[anchor=north] at (0.11\linewidth, 43mm) {
            \begin{tabular}{c@{\hspace{22mm}}c}
                HyMNet & \textbf{USCNet}  \\
            \end{tabular}   
        };
        
        \node[anchor=east, rotate=90] at (-47mm, 0.5\linewidth) {\textbf{Actual}};
        \node[anchor=east] at (-22mm, 0.45\linewidth) {
            \begin{tabular}{c}
                Non-infectious \\ \\ \\ \\Infectious \\ \\ \\ \\ \\Non-infectious \\ \\ \\ \\Infectious 
            \end{tabular}
        };
    \end{tikzpicture}
    
    \caption{Confusion matrices comparison of USCNet versus multimodal baseline methods, quantitatively evaluating classification performance for infectious urolithiasis diagnosis.}
    \label{CM}
\end{figure}

In contrast, when comparing USCNet with state-of-the-art multi-modal methods, significant advancements are clear. While methods such as HyMNet and TMSS improved Acc to 85.66\% and 74.63\%, respectively, they typically use only CT images and EHR data, integrating them through simple concatenation. USCNet, however, incorporates segmentation feature maps with multi-modal inputs using a dual self-attention module for feature fusion, markedly enhancing classification Acc. By developing a dynamic loss function, USCNet adaptively balances segmentation and classification losses based on task-specific metrics, ensuring that segmentation features effectively support classification tasks and bolstering the model's feature learning capabilities. Ultimately, USCNet achieved an Acc of 91.91\%, representing improvements of 27.94\%, 31.98\%, 30.64\%, 29.41\%, 17.28\%, and 6.25\% over ResNet architectures, StoneNet, SegPrompt, ResGANet, TMSS, and HyMNet, respectively. The performance of USCNet was further validated through DeLong's test, which confirmed a statistically significant improvement in AUC compared to the second-best method, HyMNet (\(p < 0.001\)). Figure \ref{ROC} illustrates the ROC curve, where USCNet's proximity to the top-left corner highlights its excellent performance in distinguishing between positive and negative samples. The high AUC further validates the model's Acc and reliability. Figure \ref{CM} displays the confusion matrix for USCNet, showing a relatively low number of misclassifications, thus underscoring the model's robust performance.

Beyond these quantitative metrics, USCNet provides significant clinical value by predicting stone infectivity through the integration of preoperative imaging and clinical data. This capability offers crucial decision support for perioperative management and may help reduce postoperative complication rates. Importantly, USCNet's application extends beyond high-resource clinical settings - its implementation in underserved areas could assist less experienced clinicians in making accurate diagnoses, thereby improving healthcare accessibility. This democratization of diagnostic capability is particularly valuable in regions with significant disparities in medical resources.

In the segmentation task, USCNet achieved a Dice score of 85.64\%, an IoU of 72.12\%, and an HD95 of 3.73 mm, delivering strong overall performance, second only to nnU-net. This outcome reflects the core principle of our adaptive weighting strategy: the model prioritizes extracting effective segmentation features to support classification rather than maximizing segmentation accuracy alone. Once the segmentation quality reaches a stable level sufficient to aid classification, the optimization focus shifts dynamically toward enhancing classification performance. Notably, USCNet surpassed all other comparative methods except nnU-net in HD95 boundary accuracy, indicating its segmentation masks provide more precise boundary information. By maintaining robust segmentation quality, USCNet significantly improved classification performance, effectively demonstrating the strength of our segmentation-assisted classification approach.


\begin{figure}[!tbp]
    \centering
    
    \hspace*{0.08\linewidth}\includegraphics[width=0.88\columnwidth]{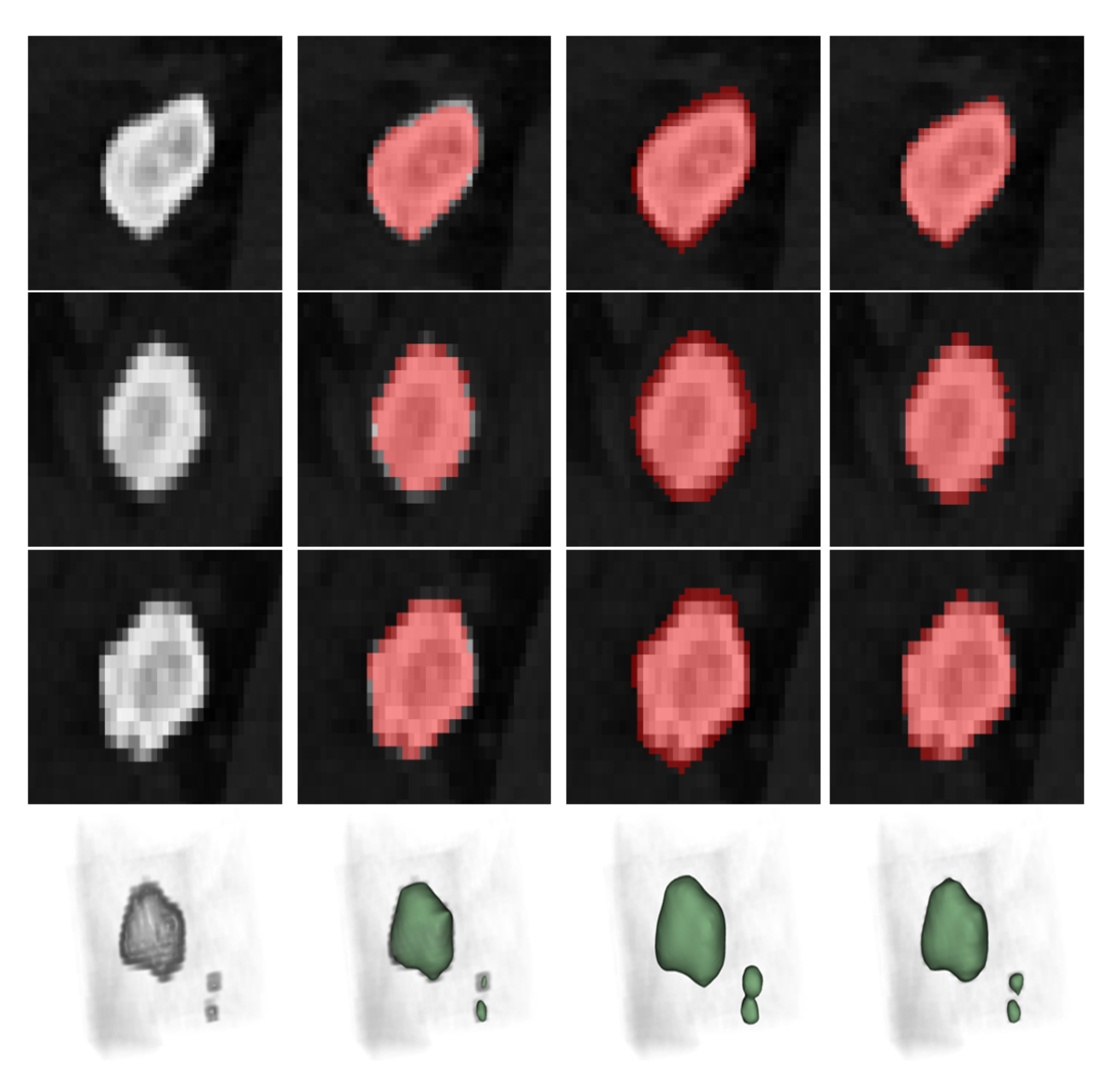}
    
    \begin{tikzpicture}[overlay]
        \node[anchor=east, rotate=0] at (-0.38\linewidth, 0.78\linewidth) {Axial};
        \node[anchor=east, rotate=0] at (-0.38\linewidth, 0.58\linewidth) {Coronal};
        \node[anchor=east, rotate=0] at (-0.38\linewidth, 0.37\linewidth) {Sagittal};
        \node[anchor=east, rotate=0] at (-0.36\linewidth, 0.17\linewidth) {3D View};
        
        \node[anchor=north] at (-0.28\linewidth, 0.05\linewidth) {Original Image};
        \node[anchor=north] at (-0.03\linewidth, 0.05\linewidth) {Ground Truth};
        \node[anchor=north] at (0.17\linewidth, 0.05\linewidth) {nnUnet};
        \node[anchor=north] at (0.37\linewidth, 0.05\linewidth) {\textbf{USCNet}};
    \end{tikzpicture}
    
    \caption{Comparison of 3D multi-planar segmentation results (original images, ground truth annotations, nnUnet and USCNet outputs).}
    \label{SEG}
\end{figure}
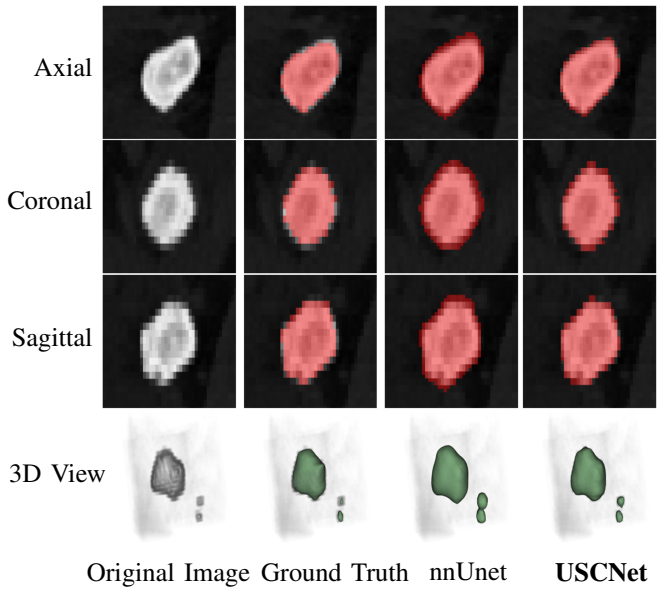

\begin{table*}[t]
\centering
\caption{Ablation studies of the proposed dynamic loss function with different components.}
\setlength{\tabcolsep}{0.8mm}
\scriptsize
\renewcommand{\arraystretch}{1.05}
\resizebox{1.0\linewidth}{!}{
\begin{tabular}{c|c|c|c|c|ccccc|cccc}
\hline
\multicolumn{5}{c|}{Loss Function Components} & \multicolumn{5}{c|}{Classification} & \multicolumn{3}{c}{Segmentation} \\ 
\cline{1-13} 
CT & Clinical & CEA & SMA & DW & Acc$\uparrow$(\%) & F1$\uparrow$(\%) & Rec$\uparrow$(\%) & Pre$\uparrow$(\%) & AUC$\uparrow$(\%) & Dice$\uparrow$(\%) & IoU$\uparrow$(\%) & HD95$\downarrow$(mm) \\
\hline
\textbf{-} & $\checkmark$ & \textbf{-} & \textbf{-} & \textbf{-} & 63.97${\pm 3.38}$ & 59.17${\pm 3.26}$ & 65.14${\pm 4.37}$ & 54.20${\pm 5.25}$ & 64.16${\pm 3.82}$ & - & - & - \\
$\checkmark$ & \textbf{-} & \textbf{-} & \textbf{-} & \textbf{-} & - & - & - & - & - & 82.27${\pm 1.02}$ & 69.18${\pm 1.25}$ & 3.88${\pm 0.51}$ \\
$\checkmark$ & $\checkmark$ & \textbf{-} & \textbf{-} & \textbf{-} & 70.56${\pm 4.88}$ & 66.90${\pm 5.62}$ & 74.33${\pm 5.27}$ & 60.90${\pm 6.15}$ & 70.16${\pm 4.84}$ & 78.33${\pm 1.73}$ & 64.30${\pm 1.73}$ & 7.42${\pm 0.87}$ \\
$\checkmark$ & $\checkmark$ & $\checkmark$ & \textbf{-} & \textbf{-} & 89.71${\pm 1.86}$ & 87.16${\pm 2.27}$ & 87.16${\pm 1.69}$ & 87.16${\pm 3.39}$ & 89.28${\pm 1.79}$ & 82.19${\pm 0.93}$ & 69.54${\pm 0.93}$ & 4.31${\pm 0.30}$ \\
$\checkmark$ & $\checkmark$ & $\checkmark$ & $\checkmark$ & \textbf{-} & 89.34${\pm 1.38}$ & 86.64${\pm 1.52}$ & 86.24${\pm 2.06}$ & 87.04${\pm 1.78}$ & 88.82${\pm 1.40}$ & 81.44${\pm 0.95}$ & 68.34${\pm 0.95}$ & 4.03${\pm 0.43}$ \\
$\checkmark$ & $\checkmark$ & $\checkmark$ & $\checkmark$ & $\checkmark$ & 
\textbf{91.91$\pm$1.42} & 
\textbf{90.00$\pm$1.74} & 
\textbf{90.83$\pm$0.94} & 
\textbf{89.19$\pm$3.13} & 
\textbf{91.73$\pm$1.30} & 
\textbf{83.98$\pm$1.09} & 
\textbf{72.12$\pm$1.42} & 
\textbf{3.73$\pm$0.82} \\
\hline
\end{tabular}}
\label{T2}
\end{table*}

\begin{table*}[t]
\centering
\caption{Ablation studies of segmentation-assisted classification with different feature combinations.}
\setlength{\tabcolsep}{0.9mm}
\scriptsize 
\renewcommand{\arraystretch}{1.05} 
\resizebox{0.95\linewidth}{!}{
\begin{tabular}{c|c|c|c|ccccc|cccc}
\hline
\multicolumn{4}{c|}{Feature Comb.} & \multicolumn{5}{c|}{Classification} & \multicolumn{3}{c}{Segmentation} \\ 
\cline{1-12} 
\(\mathbf{Z}_3\) & \(\mathbf{Z}_6\) & \(\mathbf{Z}_9\) & \(\mathbf{Z}_{12}\) & Acc$\uparrow$(\%) & F1$\uparrow$(\%) & Rec$\uparrow$(\%) & Pre$\uparrow$(\%) & AUC$\uparrow$(\%) & Dice$\uparrow$(\%) & IoU$\uparrow$(\%) & HD95$\downarrow$(mm) \\
\hline
\textbf{-} & \textbf{-} & \textbf{-} & $\checkmark$ & \textbf{91.91$\pm$1.42} & \textbf{90.00$\pm$1.74} & \textbf{90.83$\pm$0.94} & \textbf{89.19$\pm$3.13} & \textbf{91.73$\pm$1.30} & \textbf{83.98$\pm$1.09} & \textbf{72.12$\pm$1.42} & \textbf{3.73$\pm$0.82} \\
\textbf{-} & \textbf{-} & $\checkmark$ & \textbf{-} & 90.44$\pm$1.54 & 88.29$\pm$1.91 & 89.91$\pm$1.51 & 86.73$\pm$3.22 & 90.35$\pm$1.44 & 81.93$\pm$0.99 & 69.49$\pm$0.99 & 4.66$\pm$0.56 \\ 
\textbf{-} & $\checkmark$ & \textbf{-} & \textbf{-} & 90.07$\pm$2.00 & 87.78$\pm$2.34 & 88.99$\pm$2.02 & 86.61$\pm$3.50 & 89.89$\pm$1.93 & 82.29$\pm$0.96 & 69.71$\pm$0.96 & 4.18$\pm$0.49 \\ 
$\checkmark$ & \textbf{-} & \textbf{-} & \textbf{-} & 87.87$\pm$2.10 & 84.93$\pm$2.43 & 85.32$\pm$2.14 & 84.55$\pm$3.02 & 87.45$\pm$2.05 & 79.98$\pm$1.07 & 66.67$\pm$1.07 & 5.33$\pm$0.25 \\ 
\hline
$\checkmark$ & $\checkmark$ & \textbf{-} & \textbf{-} & 89.71$\pm$1.82 & 87.39$\pm$2.13 & 88.99$\pm$1.14 & 85.84$\pm$3.75 & 89.59$\pm$1.64 & 82.05$\pm$1.02 & 69.47$\pm$1.02 & 4.59$\pm$0.43 \\ 
$\checkmark$ & \textbf{-} & $\checkmark$ & \textbf{-} & 89.71$\pm$2.01 & 87.27$\pm$2.38 & 88.07$\pm$1.43 & 86.49$\pm$3.77 & 89.44$\pm$1.87 & 82.24$\pm$0.92 & 69.58$\pm$0.92 & 4.40$\pm$0.19 \\ 
$\checkmark$ & \textbf{-} & \textbf{-} & $\checkmark$ & 90.07$\pm$1.72 & 87.78$\pm$2.02 & 88.99$\pm$1.44 & 86.61$\pm$3.47 & 89.89$\pm$1.59 & 82.32$\pm$0.89 & 69.83$\pm$0.89 & 4.10$\pm$0.27 \\ 
\textbf{-} & $\checkmark$ & $\checkmark$ & \textbf{-} & 89.34$\pm$1.72 & 87.00$\pm$2.10 & 88.99$\pm$1.55 & 85.09$\pm$3.75 & 89.28$\pm$1.58 & 82.45$\pm$1.02 & 69.91$\pm$1.02 & 4.40$\pm$0.26 \\ 
\textbf{-} & $\checkmark$ & \textbf{-} & $\checkmark$ & 89.71$\pm$2.00 & 87.16$\pm$2.45 & 87.16$\pm$2.62 & 87.16$\pm$3.44 & 89.28$\pm$2.02 & 81.61$\pm$1.05 & 68.73$\pm$1.05 & 4.77$\pm$0.50 \\ 
\textbf{-} & \textbf{-} & $\checkmark$ & $\checkmark$ & 90.44$\pm$1.67 & 88.39$\pm$1.94 & 90.83$\pm$1.47 & 86.09$\pm$3.21 & 90.50$\pm$1.53 & 82.60$\pm$0.81 & 70.22$\pm$0.81 & 4.20$\pm$0.46 \\ 
\hline
$\checkmark$ & $\checkmark$ & $\checkmark$ & \textbf{-} & 89.71$\pm$1.53 & 87.27$\pm$1.69 & 88.07$\pm$1.02 & 86.49$\pm$2.67 & 89.44$\pm$1.40 & 80.80$\pm$1.08 & 67.43$\pm$1.08 & 4.32$\pm$0.39 \\ 
$\checkmark$ & $\checkmark$ & \textbf{-} & $\checkmark$ & 90.44$\pm$1.50 & 88.29$\pm$1.76 & 89.91$\pm$1.35 & 86.73$\pm$3.21 & 90.35$\pm$1.37 & 82.34$\pm$0.98 & 69.85$\pm$0.98 & 4.12$\pm$0.41 \\ 
$\checkmark$ & \textbf{-} & $\checkmark$ & $\checkmark$ & 89.34$\pm$2.16 & 86.76$\pm$2.65 & 87.16$\pm$2.09 & 86.36$\pm$3.85 & 88.98$\pm$2.10 & 82.31$\pm$0.96 & 69.85$\pm$0.96 & 4.18$\pm$0.43 \\ 
\textbf{-} & $\checkmark$ & $\checkmark$ & $\checkmark$ & 89.71$\pm$1.77 & 87.27$\pm$2.07 & 88.07$\pm$1.48 & 86.49$\pm$3.58 & 89.44$\pm$1.64 & 81.28$\pm$0.99 & 68.12$\pm$0.99 & 4.61$\pm$0.44 \\ 
\hline
$\checkmark$ & $\checkmark$ & $\checkmark$ & $\checkmark$ & 89.71$\pm$1.99 & 87.16$\pm$2.45 & 87.16$\pm$2.23 & 87.16$\pm$3.51 & 89.28$\pm$1.97 & 81.90$\pm$1.10 & 69.17$\pm$1.10 & 4.30$\pm$0.65 \\ 
\hline
\end{tabular}}
\label{T3}
\end{table*}

\begin{table*}[t]
\centering
\caption{Ablation studies of different loss functions for segmentation-assisted classification.}
\setlength{\tabcolsep}{0.9mm}
\scriptsize 
\renewcommand{\arraystretch}{1.05} 
\resizebox{0.9\linewidth}{!}{%
\begin{tabular}{c|c|c|ccccc|ccc}
\hline
\multicolumn{3}{c|}{Loss Function} & \multicolumn{5}{c|}{Classification} & \multicolumn{3}{c}{Segmentation} \\ 
\cline{1-11} 
$\mathcal{L}_{\text{BCE}}$ & $\mathcal{L}_{\text{Focal}}$ & $\mathcal{L}_{\text{Dice}}$ & Acc$\uparrow$(\%) & F1$\uparrow$(\%) & Rec$\uparrow$(\%) & Pre$\uparrow$(\%) & AUC$\uparrow$(\%) & Dice$\uparrow$(\%) & IoU$\uparrow$(\%) & HD95$\downarrow$(mm) \\
\hline
$\checkmark$ & $-$ & $\checkmark$ & 91.54$\pm$1.55 & 89.50$\pm$1.85 & 89.91$\pm$1.72 & 89.09$\pm$3.01 & 91.27$\pm$1.49 & 82.43$\pm$1.25 & 70.11$\pm$1.58 & 4.19$\pm$0.49 \\
$-$ & $\checkmark$ & $\checkmark$ & 91.18$\pm$1.20 & 88.68$\pm$1.35 & 86.24$\pm$2.38 & 91.26$\pm$1.75 & 90.36$\pm$1.30 & 83.02$\pm$2.17 & 70.96$\pm$2.32 & 4.02$\pm$1.59 \\
$\checkmark$ & $\checkmark$ & $\checkmark$ & \textbf{91.91$\pm$1.42} & \textbf{90.00$\pm$1.74} & \textbf{90.83$\pm$0.94} & \textbf{89.19$\pm$3.13} & \textbf{91.73$\pm$1.30} & \textbf{83.98$\pm$1.09} & \textbf{72.12$\pm$1.42} & \textbf{3.73$\pm$0.82} \\
\hline
\end{tabular}}
\label{T4}
\end{table*}

\begin{table*}[t]
\centering
\caption{Effect of the adaptive loss functions on the classification and segmentation performances of the USCNet model.}
\setlength{\tabcolsep}{0.9mm} 
\scriptsize 
\renewcommand{\arraystretch}{1.05} 
\resizebox{0.9\linewidth}{!}{%
\begin{tabular}{c|ccccc|cccc}
\hline
\multicolumn{1}{c|}{\multirow{2}{*}{Class. / Seg.}} & \multicolumn{5}{c|}{Classification} & \multicolumn{3}{c}{Segmentation} \\ 
\cline{2-9} 
\multicolumn{1}{c|}{} & Acc$\uparrow$(\%) & F1$\uparrow$(\%) & Rec$\uparrow$(\%) & Pre$\uparrow$(\%) & AUC$\uparrow$(\%) & Dice$\uparrow$(\%) & IoU$\uparrow$(\%) & HD95$\downarrow$(mm) \\
\hline
0.1 / 0.9 & 82.35$\pm$2.23 & 77.14$\pm$2.59 & 74.31$\pm$2.66 & 80.20$\pm$3.84 & 81.02$\pm$2.11 & 78.86$\pm$1.18 & 66.67$\pm$1.18 & 5.66$\pm$0.35 \\
0.3 / 0.7 & 86.40$\pm$1.39 & 83.41$\pm$1.36 & 85.32$\pm$1.65 & 81.58$\pm$3.18 & 86.22$\pm$1.14 & 81.96$\pm$1.11 & 69.00$\pm$1.11 & 4.38$\pm$0.64 \\
0.5 / 0.5 & 89.34$\pm$1.38 & 86.64$\pm$1.52 & 86.24$\pm$2.06 & 87.04$\pm$1.78 & 88.82$\pm$1.40 & 81.44$\pm$0.95 & 68.34$\pm$0.95 & 4.03$\pm$0.43 \\
0.7 / 0.3 & 89.71$\pm$1.90 & 87.27$\pm$2.29 & 88.07$\pm$1.46 & 86.49$\pm$3.65 & 89.44$\pm$1.78 & 82.04$\pm$1.20 & 69.25$\pm$1.20 & 4.18$\pm$0.58 \\
0.9 / 0.1 & 90.44$\pm$1.67 & 88.39$\pm$1.93 & 90.83$\pm$1.21 & 86.09$\pm$3.67 & 90.50$\pm$1.47 & 82.11$\pm$1.02 & 69.47$\pm$1.02 & 4.78$\pm$0.31 \\
Auto (Ours) & \textbf{91.91$\pm$1.42} & \textbf{90.00$\pm$1.74} & \textbf{90.83$\pm$0.94} & \textbf{89.19$\pm$3.13} & \textbf{91.73$\pm$1.30} & \textbf{83.98$\pm$1.09} & \textbf{72.12$\pm$1.42} & \textbf{3.73$\pm$0.82} \\
\hline
\end{tabular}}
\label{T5}
\end{table*}

\begin{table}[!t]
\centering
\caption{Performance analysis of USCNet across different patient subgroups.}
\setlength{\tabcolsep}{0.4mm}
\scriptsize
\renewcommand{\arraystretch}{1.05}
\resizebox{\linewidth}{!}{%
\begin{tabular}{l|ccccc|c}
\hline
\multirow{2}{*}{Subgroup} & \multicolumn{5}{c|}{Performance} & \multirow{2}{*}{N} \\ 
\cline{2-6} 
 & Acc$\uparrow$(\%) & F1$\uparrow$(\%) & Rec$\uparrow$(\%) & Pre$\uparrow$(\%) & AUC$\uparrow$(\%) & \\ 
\hline

\textbf{Age Groups} & & & & & & \\
$<$40 years & 91.80 & 87.80 & 94.74 & 81.82 & 96.99 & 61 \\
40-60 years & 90.11 & 88.00 & 86.84 & 89.19 & 94.09 & 91 \\
$>$60 years & 91.18 & 88.00 & 91.67 & 84.62 & 94.32 & 34 \\
\hline
\textbf{Gender} & & & & & & \\
Male & 90.98 & 87.91 & 93.02 & 83.33 & 95.70 & 122 \\
Female & 90.62 & 88.00 & 84.62 & 91.67 & 94.33 & 64 \\
\hline
\textbf{Stone Type} & & & & & & \\
Non-infectious & 87.58 & 84.38 & 83.08 & 85.71 & 91.52 & 161 \\
Infectious & 96.40 & 95.45 & 95.45 & 95.45 & 99.63 & 111 \\
\hline
\textbf{Stone Location} & & & & & & \\
Left Ureter & 89.53 & 85.71 & 79.41 & 93.10 & 93.04 & 86 \\
Right Ureter & 98.11 & 97.56 & 100.00 & 95.24 & 96.97 & 53 \\
Ureter & 87.30 & 87.10 & 90.00 & 84.38 & 90.00 & 63 \\
Left Kidney & 93.33 & 80.00 & 66.67 & 100.00 & 100.00 & 15 \\
Right Kidney & 88.24 & 90.91 & 83.33 & 100.00 & 100.00 & 17 \\
Kidney & 93.75 & 80.00 & 100.00 & 66.67 & 100.00 & 16 \\
Bladder & 90.00 & 90.91 & 100.00 & 83.33 & 96.00 & 10 \\
Urethra & 85.71 & 66.67 & 100.00 & 50.00 & 100.00 & 7 \\

\hline
\end{tabular}}
\label{T6}
\end{table}

\begin{figure}[!t]
\centerline{\includegraphics[width=\columnwidth]{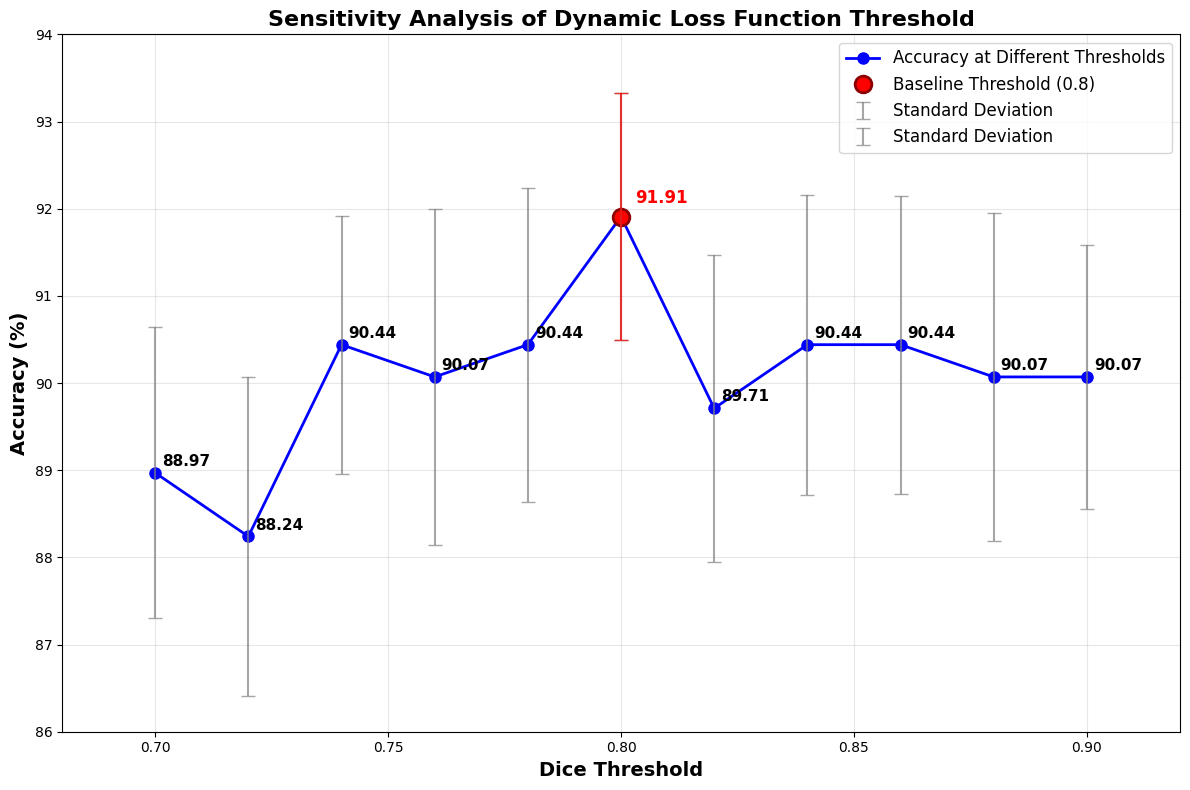}}
\caption{Empirical validation of the dynamic loss threshold through systematic sensitivity analysis. }
\label{F5}
\end{figure}

\subsection{Subgroup Analysis}

To evaluate the clinical applicability of USCNet across diverse patient populations, a detailed subgroup analysis was performed based on demographic and clinical characteristics. As summarized in Table~\ref{T6}, the model consistently demonstrated strong performance across all subgroups, with several significant trends emerging.

USCNet achieved consistently high accuracy across age groups, with results of 91.80\% for patients under 40 years, 90.11\% for those aged 40–60 years, and 91.18\% for patients over 60 years. Notably, the youngest cohort achieved the highest AUC of 96.99\%, indicating the model's ability to effectively extract features despite age-related anatomical variations. Similarly, both male and female patients showed comparable performance, with no evidence of gender-based bias affecting classification accuracy.

A particularly compelling finding was the model's exceptional performance in detecting infectious stones, achieving 96.40\% accuracy and an AUC of 99.63\%, significantly outperforming its results for non-infectious stones. This highlights USCNet's ability to identify infection-related stone characteristics, a critical factor for guiding treatment decisions such as antibiotic selection.

In terms of anatomical stone location, the model performed exceptionally well for right ureter stones and achieved perfect AUC scores for stones located in the left kidney, right kidney, and other kidney-specific locations. These results may reflect anatomical differences and distinct imaging features of stones in various segments of the urinary tract, offering valuable insights for clinical application in diverse urological settings.

\subsection{Ablation Study}
\subsubsection{Ablation Study on Key Modules}
This ablation study examines the individual contributions of key components within the proposed model, focusing on clinical data, the CEA module, the SMA module, and dynamic weight (DW) management. To evaluate the effectiveness of these components, we developed two alternative approaches: 1) excluding the CEA or SMA module, where feature fusion is done through simple data concatenation; 2) excluding the dynamic weight module, where loss weights for classification and segmentation are fixed at 0.5/0.5. These alternatives were trained alongside our original method, maintaining consistency in all other training parameters. We then evaluated each model's performance on the same test dataset using standard metrics. Detailed results are presented in Table \ref{T2}.

Initially, the baseline model, which relied solely on clinical data, achieved a classification Acc of 63.97\%. Adding CT data increased the Acc to 70.56\%, a 6.59\% improvement. However, this integration led to a decline in the Dice coefficient for segmentation, dropping from 82.27\% to 78.33\%, suggesting that handling multiple tasks might compromise segmentation performance. Introducing the CEA module, which effectively integrates multimodal data, further improved classification Acc to 89.71\%. Adding the SMA module, which leverages hidden segmentation features to enhance classification, maintained strong classification performance at 89.34\%, while segmentation Dice improved to 81.44\%.

The implementation of dynamic weight management significantly enhanced both classification and segmentation tasks. Classification Acc rose to 91.91\%, marking a 2.57\% improvement over the previous setup, while the Dice coefficient reached 83.98\% and IoU increased to 72.12\%, surpassing the performance when focusing solely on segmentation. These findings underscore the importance of each component. The CEA module initiates substantial improvements, the SMA module maintains robust classification through segmentation features, and dynamic weight management effectively balances the two tasks, maximizing the model's overall performance.

\subsubsection{Ablation Study on Hierarchical Feature Combinations}

This ablation study evaluates the influence of hidden segmentation features from various Transformer layers (\(\mathbf{Z}_3\), \(\mathbf{Z}_6\), \(\mathbf{Z}_9\), and \(\mathbf{Z}_{12}\)) on segmentation-assisted classification. As detailed in Table~\ref{T3}, leveraging only the deepest feature, \(\mathbf{Z}_{12}\), yields optimal performance across all metrics, highlighting that deeper features provide the most discriminative semantic information for both segmentation and classification tasks. 

When assessing individual shallow features, clear hierarchical differences emerge: \(\mathbf{Z}_9\) outperforms \(\mathbf{Z}_6\), while \(\mathbf{Z}_3\) yields the weakest results. This progression demonstrates that feature representational capacity increases with network depth, as shallow features primarily encode low-level spatial information. In combinations of multiple features, configurations containing \(\mathbf{Z}_{12}\) consistently deliver strong results, whereas those excluding it exhibit noticeable performance degradation. The combination of \(\mathbf{Z}_9\) and \(\mathbf{Z}_{12}\) shows effective complementarity, enhancing overall performance. However, combining all four features results in lower performance compared to the optimal single-feature setup, suggesting that excessive feature stacking introduces redundancy and noise.

In summary, the study underscores the critical role of deep features, such as \(\mathbf{Z}_{12}\), in segmentation-assisted classification. It emphasizes the importance of strategic feature selection over indiscriminate stacking to achieve optimal multimodal integration and maximize overall performance.

\subsubsection{Ablation Study on Loss Components}

To further investigate the impact of different loss functions on the performance of segmentation-assisted classification, we conducted an ablation study with three combinations: $\mathcal{L}_{\text{BCE}}+\mathcal{L}_{\text{Dice}}$, $\mathcal{L}_{\text{Focal}}+\mathcal{L}_{\text{Dice}}$, and $\mathcal{L}_{\text{BCE}}+\mathcal{L}_{\text{Focal}}+\mathcal{L}_{\text{Dice}}$, as shown in Table~\ref{T4}. When a loss term was absent, its weight was reallocated to the remaining classification loss to ensure fair comparisons.

The results demonstrate that the full combination of $\mathcal{L}_{\text{BCE}}$, $\mathcal{L}_{\text{Focal}}$, and $\mathcal{L}_{\text{Dice}}$ yields the best performance across both classification and segmentation tasks. Specifically, $\mathcal{L}_{\text{BCE}}$ provides stable gradient signals, $\mathcal{L}_{\text{Focal}}$ focuses on hard-to-classify examples, and $\mathcal{L}_{\text{Dice}}$ ensures accurate region-level supervision. In contrast, the two-loss combinations show inferior performance, suggesting that missing any one component weakens the model's ability to optimize both tasks jointly. These findings highlight that a well-balanced integration of complementary loss functions is critical for achieving robust and comprehensive performance in multitask learning.

\subsection{Comparative Study of Adaptive Loss Functions}

\subsubsection{Performance Comparison: Fixed and Adaptive Methods}
This study explores the advantages of adaptive loss functions and compares their effectiveness across various tasks. Through a series of experiments, we assessed how adaptive loss functions can enhance model performance compared to fixed-weight configurations, with the results detailed in Table \ref{T5}.

Different fixed-weight settings had varying impacts on performance. For example, a weight ratio of 0.5/0.5 yielded a classification Acc of 89.34\%, an F1 of 86.64\%, and a Dice coefficient of 81.44\%, showing competitive performance. This suggests that while a balanced fixed-weight approach can sustain strong performance, it lacks the nuanced adaptability that automatic adjustment methods offer. Other configurations, such as weight ratios of 0.7/0.3 and 0.9/0.1, prioritized classification Acc, achieving 89.71\% and 90.44\%, respectively. However, this emphasis on classification led to compromises in segmentation quality, with HD95 values increasing to 4.18mm and 4.78mm, illustrating the trade-off between classification performance and segmentation accuracy when segmentation is deprioritized.

These findings highlight the flexibility of adaptive loss functions, which dynamically adjust task-specific weight distributions, thereby overcoming the limitations of fixed-weight settings. By implementing adaptive loss functions with automatic weight adjustment, we observed significant improvements in both classification and segmentation tasks. Classification Acc reached 91.91\%, and the segmentation performance achieved a Dice coefficient of 83.98\%, IoU of 72.12\%, and HD95 of 3.73mm. This substantial enhancement underscores the efficacy of the adaptive approach in achieving optimal performance across tasks, demonstrating its ability to adjust loss weights according to task-specific needs, leading to superior multi-task learning outcomes.

\subsubsection{Empirical Validation of Dynamic Loss Threshold}

This study examines the impact of threshold selection for the dynamic loss function through sensitivity analysis. As illustrated in Figure~\ref{F5}, classification accuracy was evaluated across various Dice thresholds ranging from 0.7 to 0.9. The analysis revealed that a threshold of 0.8 achieved the highest accuracy of 91.91\%, outperforming all other values. Thresholds below 0.78 resulted in decreased accuracy, while those above 0.8 showed no additional improvement.

These results confirm that 0.8 is the optimal threshold for balancing segmentation and classification tasks within the multi-task framework. By ensuring a smooth transition between segmentation-focused and classification-focused training, the selected threshold contributes significantly to the overall performance enhancements demonstrated in this study.


\subsection{Discussions and Limitations}\label{DIS}
\subsubsection{Advantages and Innovations}
The USCNet framework makes three fundamental contributions to multimodal medical image analysis. First, its unified architecture combining CT imaging and EHR data establishes a new paradigm for segmentation-assisted classification in Urolithiasis management. Second, the novel cross-attention mechanism provides interpretable fusion of radiographic and clinical features through learnable projection matrices, outperforming traditional concatenation approaches. Third, the dynamic loss function implements curriculum learning by automatically shifting focus from segmentation to classification tasks during training, achieving superior optimization balance compared to fixed-weight schemes.

In addition to its technical advantages, USCNet offers distinct clinical benefits. Traditional methods require postoperative specimen images as input, limiting their practical clinical use. Conversely, our model innovatively utilizes preoperative imaging and clinical data to anticipate the infectivity of stones, providing essential support for decision-making during perioperative management and potentially minimizing postoperative complications. The predictive outcomes play a pivotal role in guiding clinical decision-making. For infectious stones, the model enables clinicians to initiate targeted antibiotic therapy prior to surgery and refine surgical strategies accordingly. Conversely, for non-infectious stones, it helps avoid unnecessary antibiotic use, allowing a greater focus on metabolic management and tailored treatment approaches. Importantly, USCNet is accessible not only to well-equipped medical institutions but also to those with limited resources. This adaptability allows physicians at various experience levels to make accurate diagnoses, thereby enhancing equitable access to high-quality urological care.

\subsubsection{Limitations and Future Directions}
While USCNet offers several advantages, there are notable limitations that must be addressed. First, the model's performance heavily relies on the completeness and quality of multimodal data, particularly the accuracy and availability of laboratory test results in EHR. Second, its generalization capability requires improvement in handling complex clinical scenarios, such as cases with anatomical variations or multiple stone types. Third, models developed using single-center data are prone to spectrum bias, which may limit their applicability to datasets from other medical institutions. Fourth, the computational complexity of the Transformer architecture constrains its use in resource-limited environments. Future research will aim to address these challenges by focusing on the development of lightweight network architectures to enhance computational efficiency. Additionally, multi-center validation studies will be conducted to better assess and improve the model's generalization capabilities. Efforts will also be directed toward expanding clinical applications, including real-time stone analysis and surgical navigation, to broaden the practical utility of the model.

\section{Conclusion}

\label{Conclusion}
This study proposed a novel multi-task USCNet model for precise segmentation and classification of urinary stones in preoperative settings. This method achieved effective integration of CT images and electronic health records through a dedicated CT-EHR attention module, while employing an SMA module that significantly improves feature extraction accuracy compared to conventional approaches. The introduced dynamic loss function effectively balanced multi-task learning with accelerated convergence. Experimental results demonstrated superior performance compared to existing methods on our proprietary kidney stone dataset across multiple classification metrics. However, the model exhibited certain limitations, including a strong dependence on the quality of multimodal data, limited generalizability in complex clinical scenarios, and relatively high computational complexity. Future research will focus on designing a lightweight architecture for computational efficiency and extending the model's applications to broader clinical settings, including real-time stone analysis and surgical planning assistance.

\section{Compliance With Ethical Standards}
This retrospective study was approved by the institutional review board of Longgang District People’s Hospital of Shenzhen (2022065).

\section*{References}
\vspace{-2em}
\bibliographystyle{IEEEtran}
\bibliography{ref}
\end{document}